\documentclass[10pt,twocolumn,letterpaper]{article}

\usepackage[cvpr]{cvpr}
\usepackage{algorithm}
\usepackage{algorithmic}
\usepackage{multirow}
\usepackage{listings}
\usepackage[most]{tcolorbox}
\usepackage{xcolor}

\definecolor{cvprblue}{rgb}{0.21,0.49,0.74}
\usepackage[pagebackref,breaklinks,colorlinks,allcolors=cvprblue]{hyperref}

\def\paperID{*****} 
\def\confName{CVPR}
\def\confYear{2025}

\title{ Prmpt2Adpt: Prompt-Based Zero-Shot Domain Adaptation for Resource-Constrained Environments}

\author{
Yasir Ali Farrukh \quad Syed Wali \quad Irfan Khan\\
Clean and Resilient Energy System Lab (CARES), Texas A\&M University\\
College Station, TX, USA\\
{\tt\small \{yasir.ali, syedwali, irfankhan\}@tamu.edu}
\and
Nathaniel D. Bastian\\
Robotics Research Center, United States Military Academy\\
West Point, NY, USA\\
{\tt\small nathaniel.bastian@westpoint.edu}
}

\begin{document}
\maketitle

\begin{abstract}
Unsupervised Domain Adaptation (UDA) is a critical challenge in real-world vision systems, especially in resource-constrained environments like drones, where memory and computation are limited. Existing prompt-driven UDA methods typically rely on large vision-language models and require full access to source-domain data during adaptation, limiting their applicability. In this work, we propose Prmpt2Adpt, a lightweight and efficient zero-shot domain adaptation framework built around a teacher-student paradigm guided by prompt-based feature alignment. At the core of our method is a distilled and fine-tuned CLIP model, used as the frozen backbone of a Faster R-CNN teacher. A small set of low-level source features is aligned to the target domain semantics—specified only through a natural language prompt—via Prompt-driven Instance Normalization (PIN). These semantically steered features are used to briefly fine-tune the detection head of the teacher model. The adapted teacher then generates high-quality pseudo-labels, which guide the on-the-fly adaptation of a compact student model. Experiments on the MDS-A dataset demonstrate that Prmpt2Adpt achieves competitive detection performance compared to state-of-the-art methods, while delivering up to 7× faster adaptation and 5× faster inference speed using few source images—making it a practical and scalable solution for real-time adaptation in low-resource domains.
\end{abstract}    
\section{Introduction}
\label{sec:intro}
Domain shift, referring to the disparity between the training data distribution (source domain) and the testing data distribution (target domain), continues to be a fundamental challenge in machine learning \cite{areo2025overcoming}. Deep models trained under fixed conditions often degrade significantly when deployed in unseen environments \cite{vance2024measuring}. This problem is amplified in resource-constrained settings such as drones, where limitations on compute, memory, and real-time inference capability hinder adaptability \cite{cereda2024training}.

Most domain adaptation (DA) methods assume access to unlabeled target-domain data during training, but this is rarely feasible in real-world deployments where future domains are unknown or inaccessible \cite{kouw2019review}. Addressing zero-shot adaptation—without access to target-domain images—is essential for developing models that generalize reliably to dynamic, unseen conditions.

\begin{figure}[!th]
  \centering
  \includegraphics[width=\linewidth]{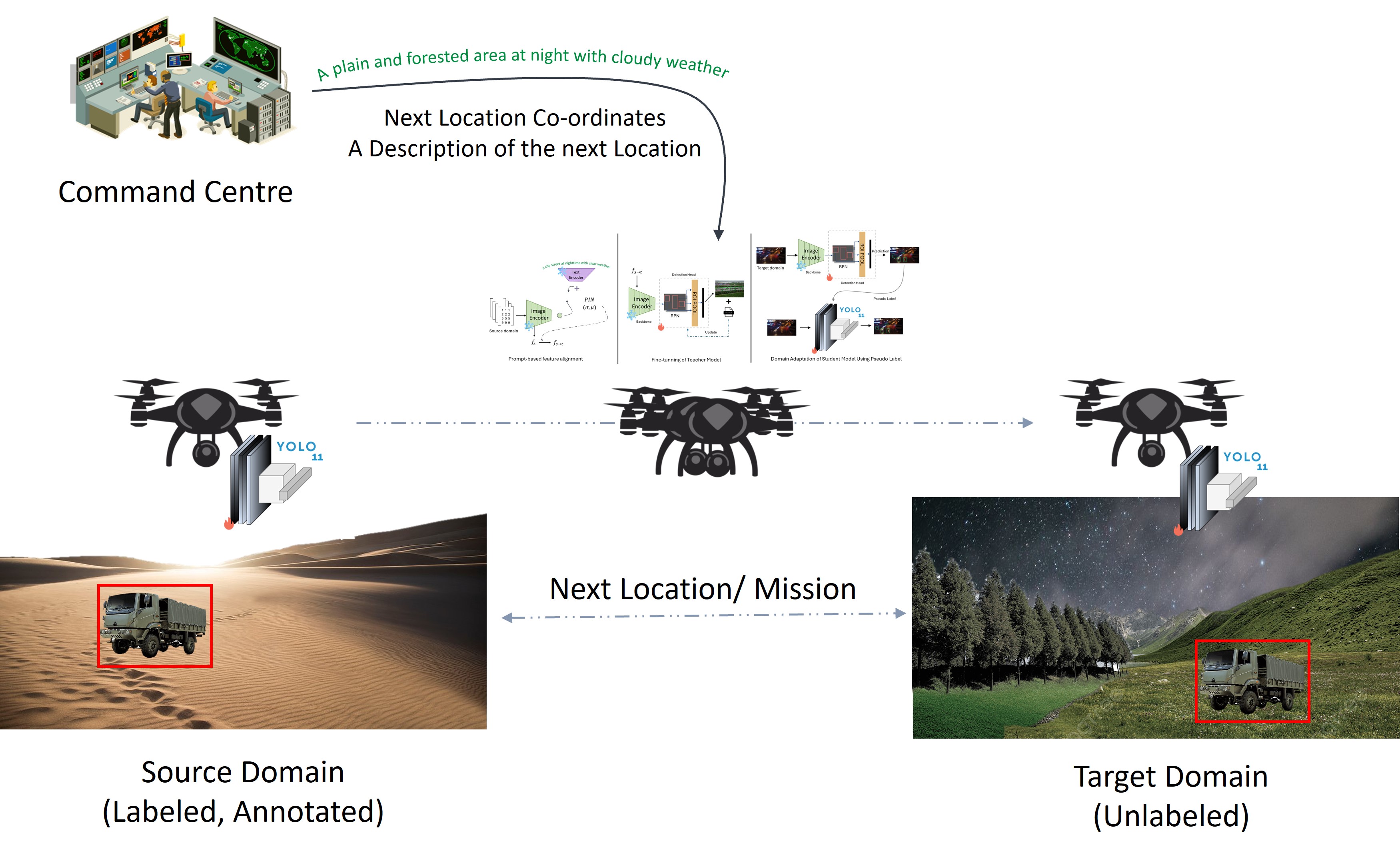}
   \caption{Prmpt2Adpt addresses UDA in a zero-shot setting, where the model is trained solely on the source domain and adapts to the target domain using only textual prompts. The goal is to enable a lightweight, fast object detection model that can adapt on-the-fly without requiring target domain images.}

  \label{fig:Scenario}
\end{figure}

Recent advances in vision-language models like CLIP \cite{clip} have enabled semantic alignment through image-text embeddings, opening up new possibilities for prompt-driven adaptation \cite{ge2023domain,kwon2022clipstyler}. However, these approaches typically rely on large models and high computational overhead, making them unsuitable for edge environments such as UAVs.

To address this gap, we propose Prmpt2Adpt, a lightweight, prompt-based zero-shot domain adaptation framework tailored for resource-constrained deployments. By leveraging natural language prompts as semantic anchors, our method adapts models to new domains without requiring target-domain images. As shown in Figure~\ref{fig:Scenario}, our system integrates efficient feature steering, knowledge transfer through a teacher-student paradigm, and a novel captioning pipeline to build a text-aligned dataset for aerial imagery.

\begin{itemize}
  \item We introduce Prmpt2Adpt framework, which leverages textual prompts as semantic anchors to steer source-domain features toward target-domain distributions—enabling adaptation without direct access to target domain data.
  \item We develop a captioning pipeline using LLaMA 3.2 to generate scenario-specific textual descriptions, forming the basis for a text-driven dataset tailored for domain adaptation.
  \item We fine-tune a distilled CLIP model on aerial datasets to enhance domain generalization and construct a lightweight teacher-student architecture that enables fast, efficient adaptation.
\end{itemize}

\section{Related Work}
\label{sec:Related_work}
\subsection{Domain Adaptation for Object Detection.} 
Numerous approaches have been proposed to tackle domain shift in object detection by aligning the source and target domain distributions. These methods generally focus on bridging gaps at global or instance levels. A prominent direction involves adversarial training, which facilitates feature alignment across domains \cite{chen2020harmonizing,saito2019strong,shen2019scl}. Other techniques leverage category-level centroids \cite{zhu2019adapting} or utilize attention maps \cite{vs2021mega} to enhance instance-level alignment specifically. Moreover, generating pseudo-labels directly from target domain images to guide target-aware training has also emerged as an effective strategy \cite{deng2021unbiased,li2022cross}.

\subsection{Unsupervised Domain Adaptation (UDA)}

Unsupervised Domain Adaptation (UDA) has been extensively explored, encompassing a variety of approaches that address different aspects of the domain shift. Methods such as adversarial learning \cite{tsai2018learning,ganin2016domain}, self-training \cite{zou2019confidence, li2019bidirectional}, entropy minimization \cite{vu2019advent,pan2020unsupervised}, and generative-based adaptation \cite{hoffman2018cycada} have been utilized to mitigate the domain gap. These approaches typically operate at different levels: input \cite{hoffman2018cycada,yang2020fda}, features  \cite{long2018conditional,ganin2016domain,wang2017deep}, or output \cite{pan2020unsupervised,vu2019advent}. Despite their progress, these methods still struggle when the target domain deviates significantly from the source domain.

One-Shot Unsupervised Domain Adaptation (OSUDA) \cite{wan2020one} represents a particularly challenging setting compared to conventional UDA, as only a single unlabeled target-domain image is available for adaptation, rather than full access to multiple unlabeled target images. Luo et al. \cite{luo2020adversarial} demonstrated that traditional UDA methods struggle under these restrictive conditions and proposed a style-mining algorithm to avoid overfitting the model to the single target image’s style.  Similarly, Wu et al. \cite{wu2022style} proposed style mixing coupled with patch-wise prototypical matching (SM-PPM), which facilitates domain adaptation by linearly combining channel-wise statistics from source and target features. 

Extending adaptation challenges further, the zero-shot domain adaptation setting \cite{du2024boosting} involves adapting a model without any access to target-domain images. Lengyel et al. \cite{lengyel2021zero} addressed such a scenario specifically for day-to-night adaptation by introducing a color-invariant convolution layer (CIConv), making neural networks robust against variations in lighting conditions. In contrast, our framework tackles the zero-shot adaptation scenario differently, leveraging textual descriptions to guide adaptation toward target-domain semantics, without direct exposure to target-domain images.

\subsection{Text-Driven Domain Adaptation}

Recent advancements in contrastive image-language pretraining, particularly CLIP, have revolutionized multimodal learning and yielded impressive results in tasks like zero-shot classification \cite{radford2021learning}, multi-modal retrieval \cite{jia2021scaling}, and visual question answering \cite{li2021align}. These breakthroughs have also enabled the modification of images using text descriptions, a task previously hindered by the disconnect between vision and language representations. For instance, StyleCLIP \cite{patashnik2021styleclip} uses prompts to optimize the latent vectors of StyleGAN \cite{karras2019style}, guiding the generation process. However, this generation is restricted to the training distribution of StyleGAN. StyleGAN-NADA \cite{gal2022stylegan} overcomes this limitation by utilizing CLIP embeddings of text-prompts to perform domain adaptation on the generator, which is trainable. Similarly, FlexIT \cite{couairon2022flexit} uses text-guided semantic image editing by optimizing the latent code in VQGAN’s \cite{esser2021taming} space. Another related work, CLIPstyler \cite{kwon2022clipstyler}, facilitates text-guided style transfer without relying on a generative process, providing more flexibility by not being restricted to specific distributions. Despite the promising results in text-guided image synthesis and editing, these works focus primarily on image classification or stylization tasks.

The closest works to ours in the domain of zero-shot adaptation is \cite{fahes2023poda}, which propose a prompt-driven zero-shot domain adaptation paradigm in computer vision, utilizing natural language descriptions of the target domain without needing target-domain images for training. Similarly, the ULDA framework \cite{yang2024unified} enables a single model to adapt to diverse target domains using only textual descriptions, eliminating the need for domain-specific labels or separate models. While these methods are closely related to our task, they primarily focus on segmentation and rely heavily on large vision-language models (VLMs) and complex fine-tuning pipelines, which involve access to the entire source domain dataset during adaptation. This presents a significant limitation in practical environments, particularly for resource-constrained scenarios, where carrying large datasets is infeasible. In contrast, our framework addresses these challenges by focusing on resource efficiency, ensuring that our approach is faster, more compact, and suitable for real-time inference, even in environments with limited computational resources.

\section{Method}
Prmpt2Adpt tackles the challenging scenario of zero-shot UDA specifically tailored for resource-constrained environments, such as drones. Existing state-of-the-art approaches for UDA typically rely heavily on large-scale VLMs and computationally expensive training pipelines. Such methods usually necessitate the storage of extensive source-domain datasets for adaptation, making them impractical for real-time inference and deployment on devices with limited computational resources and memory capacity.

In contrast to these resource-intensive methods, we propose Prmpt2Adpt, a prompt-based zero-shot domain adaptation framework specifically designed to address efficiency constraints. Our method emphasizes rapid and lightweight domain adaptation guided by textual prompts without requiring direct access to extensive datasets during adaptation, thereby making it highly suitable for real-world deployment in drone-based applications.
\subsection{Problem Formulation}
Given a labeled source-domain dataset $\mathcal{D}^{\text{Source}} = \{(X_i^{\text{Source}}, Y_i^{\text{Source}})\}_{i=1}^{N_s}$, where $X_i^{\text{Source}} \in \mathbb{R}^{H \times W \times 3}$ represents input images and $Y_i^{\text{Source}} \in \mathcal{Y}_{det}$ denotes corresponding annotations such as bounding boxes and category labels. Conversely, the target-domain dataset $\mathcal{D}^{\text{Target}} = \{X_j^{\text{Target}}\}_{j=1}^{N_t}$, with images $X_j^{\text{Target}} \in \mathbb{R}^{H \times W \times 3}$, is completely unlabeled, unseen, and unavailable during the training phase. However, auxiliary textual descriptions characterizing the semantic contexts of the source and target domains, $T^{\text{Source}}$ and $T^{\text{Target}}$, respectively, are accessible and utilized as semantic guidance for adaptation.

The primary objective is to adapt a lightweight student model $M_{\text{Student}}$, initially trained exclusively on the labeled source domain $\mathcal{D}^{\text{Source}}$, so it generalizes effectively to the unseen and unlabeled target domain $\mathcal{D}^{\text{Target}}$. During deployment, the compact student model $M_{\text{Student}}$ leverages pseudo-labels generated by a resource-aware teacher model $M_{\text{Teacher}}$. These pseudo-labels are obtained by aligning source-domain features with the textual description $T^{\text{Target}}$, allowing the student model to rapidly and effectively adapt to the target domain. This teacher-student adaptation paradigm enables efficient, on-the-fly domain adaptation without incurring additional computational or storage overhead. It is important to emphasize that the ground-truth annotations for the target-domain data, $Y^{\text{Target}}$, remain inaccessible during the adaptation phase and are strictly used for evaluation purposes.

\begin{figure}[!th]
  \centering
  \includegraphics[scale=0.4]{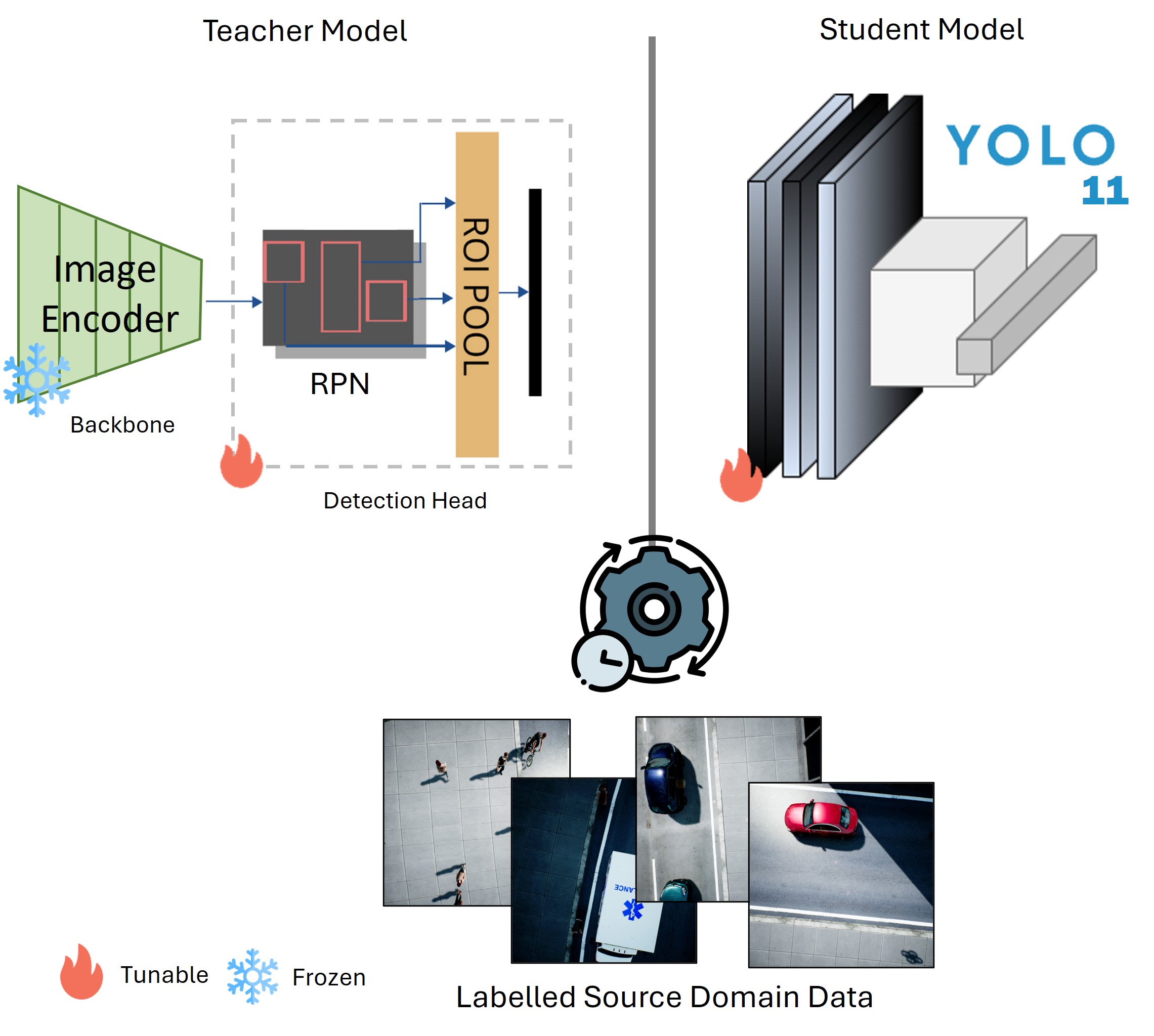}
   \caption{Overview of the teacher-student model architectures used in our proposed framework. The teacher model employs a distilled CLIP backbone (image encoder) with frozen weights, combined with a detection head comprising a Region Proposal Network (RPN) and ROI pooling layers, trained specifically on labeled source-domain data. The student model utilizes the lightweight YOLOv11 nano architecture, effective for resource-constrained environments.}

  \label{fig:Models}
\end{figure}

\begin{figure*}[!th]
  \centering
  \includegraphics[width=\linewidth]{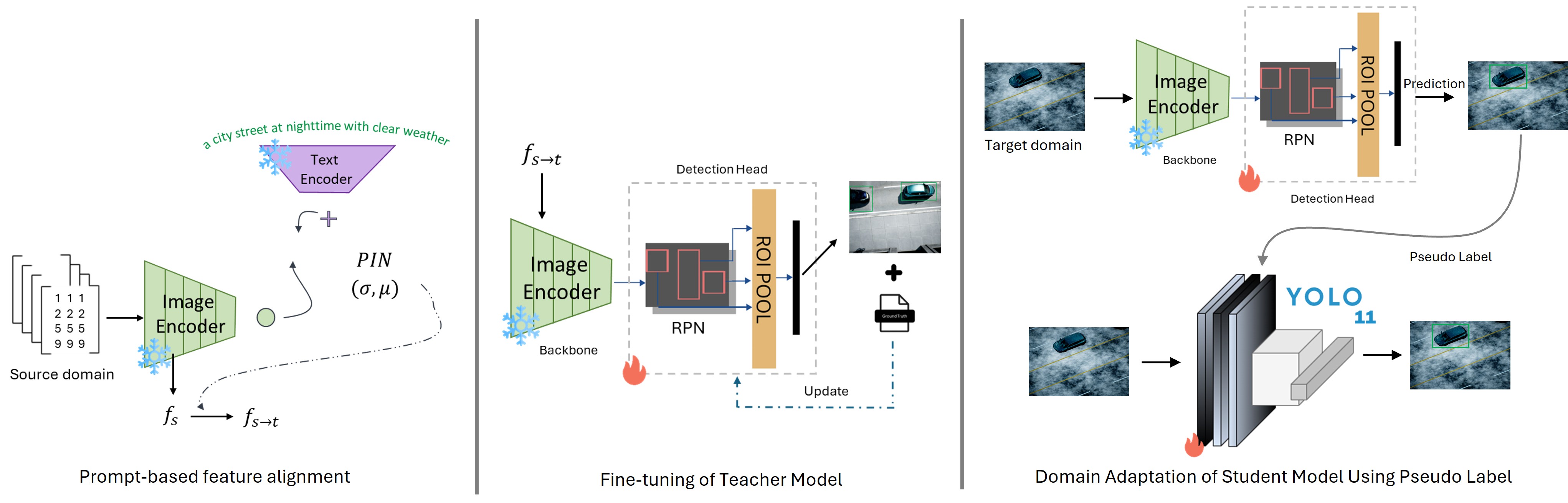}
   \caption{Overview of our proposed Prmpt2Adpt framework. The left panel illustrates the prompt-based feature alignment, where embeddings from a distilled CLIP model's image and text encoders guide the steering of source-domain features (\( f_s \)) toward the target-domain distribution (\( f_{s \rightarrow t} \)), by optimizing statistical distribution parameters (\(\sigma, \mu\)). The central panel depicts the fine-tuning process of the teacher model's detection head using these steered features along with the original source-domain ground truth. Finally, the right panel demonstrates the adaptation of the lightweight YOLOv11 nano student model using pseudo-labels generated by the adapted teacher model, enabling efficient real-time performance in the target domain without additional computational overhead.}

  \label{fig:Approach}
\end{figure*}

\subsection{Prmpt2Adpt Overview}
Our proposed framework, Prmpt2Adpt, introduces a lightweight and efficient teacher-student paradigm, incorporating the feature-steering principle inspired by \cite{fahes2023poda}, specifically designed for robust zero-shot unsupervised domain adaptation (UDA) in resource-constrained environments. Such settings, particularly drone-based applications, impose strict limitations on computational resources, memory, and real-time inference capabilities, making conventional adaptation techniques impractical.

To address these constraints, we employ two complementary models optimized explicitly for resource efficiency. The student model utilizes the compact YOLOv11 nano architecture \cite{yolo11_ultralytics}, renowned for its real-time and efficient object detection capabilities. Despite these advantages, YOLOv11 nano often faces considerable performance degradation when encountering significant domain shifts common in drone scenarios. Therefore, to facilitate robust adaptation, we introduce a teacher model based on the Faster R-CNN detection framework. This teacher model integrates a distilled CLIP vision encoder backbone with a specialized detection head comprising a Region Proposal Network (RPN) and Region-of-Interest (ROI) pooling layers. Figure \ref{fig:Models} visually illustrates the architectures of both the teacher and student models, highlighting their respective trainable parameters and datasets used for training.

The choice of the distilled CLIP backbone is motivated by its rich semantic embeddings obtained via contrastive language-image pre-training. While original CLIP models, typically transformer-based, demand substantial computational resources, our distilled CLIP backbone significantly reduces complexity while preserving critical semantic representation capabilities. Importantly, the distilled CLIP backbone's weights are frozen during adaptation, ensuring stability of semantic embeddings, thereby facilitating efficient feature steering guided solely by textual prompts.

The teacher model, benefiting from the frozen distilled CLIP backbone, produces rich image representations aligned closely with textual descriptions. Training its detection head on source-domain datasets aligned with typical drone-based scenarios (as depicted in Figure \ref{fig:Scenario}), enables our teacher model to effectively capture visual-semantic relationships essential for domain adaptation. Consequently, this teacher model guides the adaptation by steering source-domain features toward the target-domain semantic context described by textual prompts.

The high-level overview of our proposed framework is depicted in Figure \ref{fig:Approach}. In the left panel, we illustrate the distilled CLIP backbone producing embeddings for both the source-domain image and the provided textual prompt. By optimizing the statistical parameters (mean \(\mu\) and standard deviation \(\sigma\)) of the source image features via a contrastive loss between these embeddings, we effectively steer the source-domain features toward the target-domain distribution guided by the textual semantic description. The central panel of Figure \ref{fig:Approach} demonstrates how these steered and augmented features, alongside the original source-domain ground truth labels, are utilized to refine and update the detection head of the teacher model, significantly enhancing its ability to generalize to previously unseen target domains. Finally, the rightmost panel of Figure \ref{fig:Approach} shows how, upon completion of this adaptation step, the enhanced teacher model generates accurate pseudo-labels for the lightweight student YOLOv11 nano model. Utilizing these pseudo-labels, the student model rapidly and efficiently adapts during the early stages of target-domain deployment, substantially improving its detection performance without incurring additional computational or memory overhead.

\subsection{Teacher and Student Model}
\textbf{Teacher Model:} Our teacher model leverages a distilled version of the CLIP image encoder to ensure effective representation learning while maintaining efficiency for resource-constrained environments. Specifically, we adopt the TinyCLIP framework to distill the original CLIP ResNet-50 model into a significantly smaller variant, TinyCLIP (ResNet-19M, Text-19M). This distillation approach, inspired by Wu et al. \citep{tinyclip}, involves affinity mimicking—where the distilled model replicates the image-text alignment behavior by optimizing similarity distributions across both image-to-text and text-to-image affinities. Additionally, weight inheritance is employed, wherein we initialize the distilled model using a carefully selected subset of the original CLIP model's weights. This step involves manually reducing the width of the vision encoder and depth of the text encoder to preserve representational power while significantly reducing computational complexity. Finally, a multi-stage progressive distillation strategy is adopted to gradually compress the model, ensuring stable convergence and retention of cross-modal alignment capability. Through this structured distillation process, we achieve a compact and efficient CLIP model approximately half the size and twice the inference speed of the original, making it particularly suitable for resource-constrained environments.

\begin{figure}[!th]
  \centering
  \includegraphics[width=\linewidth]{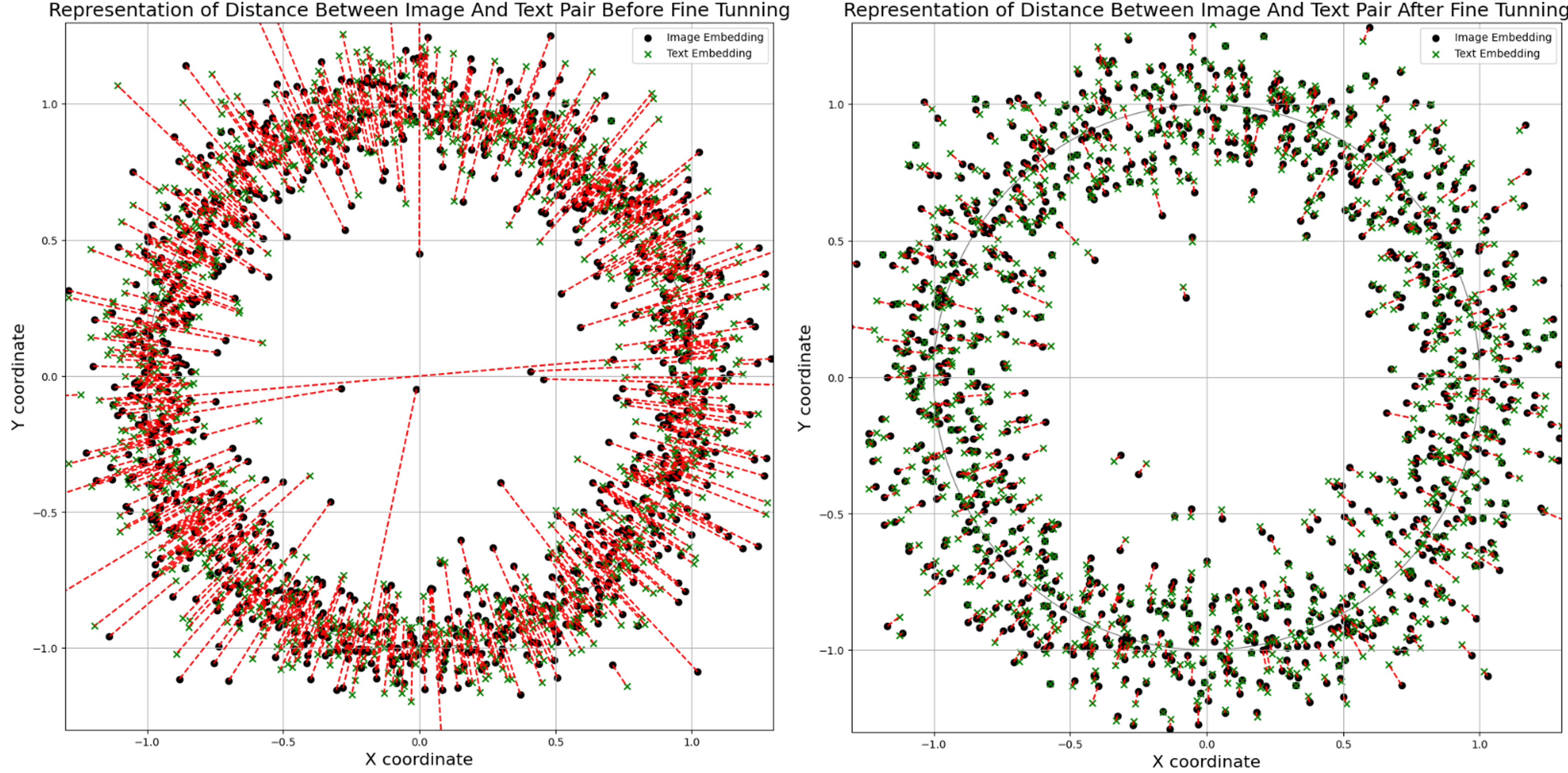}
   \caption{Visualization of embedding distances between image-text pairs before and after fine-tuning the distilled CLIP model. The left plot illustrates significant embedding distances prior to fine-tuning, indicating poor semantic alignment. After fine-tuning (right plot), the distances between corresponding image-text pairs are considerably reduced, demonstrating improved semantic representation and enhanced alignment for scenario-based descriptions.}

  \label{fig:embd_dist}
\end{figure}

Despite these advantages, the pretrained distilled CLIP model initially demonstrated poor alignment between image-text pairs in our specific scenario-based problem setting. During testing, image-text pairs describing scenarios and backgrounds exhibited low similarity scores and larger embedding distances, indicating inadequate representational capability. To demonstrate embedding distances and similarity, we plotted the image and text embeddings using an around-unit-circle visualization, highlighting significant embedding distances prior to fine-tuning, as illustrated in Figure \ref{fig:embd_dist}. To overcome this limitation, we explicitly fine-tuned our distilled CLIP model on UAVDT \cite{uavdt}, AID \cite{AID}, VDD \cite{VDD}, and NAT2021 \cite{NAT2021} datasets. These datasets contain diverse drone imagery scenarios and corresponding textual descriptions generated via our automated captioning pipeline. Since the original CLIP model was primarily trained for object classification, scenario-based textual prompts lacked robust semantic representation. Fine-tuning substantially improved the semantic alignment between images and their textual descriptions, as evidenced by the reduced embedding distances in our unit-circle plots.

After fine-tuning, we froze the distilled CLIP model weights to maintain compatibility with the original CLIP latent space and ensure consistent semantic representations. Additionally, we removed the attention pooling head from the vision encoder to retain spatial information crucial for object detection. With this refined encoder, we constructed our teacher detection model by integrating the frozen distilled CLIP vision encoder as a backbone with a detection head composed of a RPN and ROI pooling layers. Thus, only the detection head parameters remain trainable during domain adaptation, efficiently leveraging the rich visual-semantic embeddings from the frozen distilled CLIP backbone.

\textbf{Student Model:} For our student model, we adopt the lightweight YOLOv11 nano architecture \cite{yolo11_ultralytics}, renowned for its efficient real-time inference capabilities. Despite these computational advantages, YOLOv11 nano can experience accuracy degradation when subjected to domain shifts. To address this challenge, the student model initially leverages pseudo-labels generated by the teacher model during the early stages of inference, enabling rapid and effective domain adaptation. Subsequently, the adapted student model independently performs inference. This teacher-student paradigm allows YOLOv11 nano to adapt seamlessly to new domains without additional computational overhead, ensuring robust and efficient real-time detection performance.

\subsection{Prompt-based Feature Alignment}

To align the distribution of source-domain features with a target domain defined through a natural language prompt, we leverage the dual-encoder architecture of CLIP to implement a prompt-guided feature alignment mechanism. Specifically, we extract low-level visual features $f_s \in \mathbb{R}^{h \times w \times c}$ from the Layer1 of the frozen distilled CLIP image encoder. These extracted features capture essential visual information necessary for successful domain adaptation, preserving semantic and spatial details.

To steer these source-domain features toward the semantics of the target domain described by the textual prompt, we employ Prompt-driven Instance Normalization (PIN)~\cite{fahes2023poda}. PIN is inspired by Adaptive Instance Normalization (AdaIN)~\cite{huang2017arbitrary} but explicitly guided by CLIP's latent embedding space. Formally, given a low-level feature map $f_s$, PIN applies an affine transformation controlled by optimizable style statistics $(\mu, \sigma)$, defined as:

\begin{equation}
    \text{PIN}(f_s, \mu, \sigma) = \sigma \cdot \left( \frac{f_s - \mu(f_s)}{\sigma(f_s)} \right) + \mu,
\end{equation}

where $\mu(f_s)$ and $\sigma(f_s)$ represent the channel-wise mean and standard deviation of the original source features, respectively. The parameters $(\mu, \sigma)$ are optimized to align the embedding $\bar{f}_{s \rightarrow t}$ of the transformed features $f_{s \rightarrow t}$ with the target prompt embedding $\text{TrgEmb} = E_{\text{txt}}(T_{\text{prompt}})$ in CLIP's joint vision-language embedding space. This optimization is achieved by minimizing the cosine distance loss:

\begin{equation}
    \mathcal{L}_{\mu, \sigma} = 1 - \frac{\bar{f}_{s \rightarrow t} \cdot \text{TrgEmb}}{\|\bar{f}_{s \rightarrow t}\| \cdot \|\text{TrgEmb}\|}.
\end{equation}

This optimization effectively guides the source-domain features toward the semantic characteristics of the target prompt while preserving their inherent spatial and semantic content. Consequently, it enables efficient zero-shot domain adaptation purely driven by textual prompts without direct exposure to target-domain images. A detailed summary of the prompt-based feature alignment procedure is provided in Algorithm~\ref{algo:pin}.

\begin{algorithm}[t]
\caption{Prompt-based Feature Alignment}
\label{algo:pin}
\begin{algorithmic}[1]
\REQUIRE $\mathcal{F}_s$: Set of source features\\
\hspace*{2.3em} $\text{TrgEmb}$: Target prompt embedding  \\
\hspace*{2.3em} $N$: Number of optimization steps $N$ \\
\hspace*{2.3em} $lr$: Learning rate $lr$ \\
\hspace*{2.3em} $m$: Momentum

\STATE Initialize $\mathcal{S}_{s \rightarrow t} \leftarrow \emptyset$
\FORALL{$f_s \in \mathcal{F}_s$}
    \STATE $\mu^0 \leftarrow \text{mean}(f_s)$
    \STATE $\sigma^0 \leftarrow \text{std}(f_s)$

    \FOR{$i = 1$ to $N$}
        \STATE $f_{s \rightarrow t}^{i} \leftarrow \text{PIN}(f_s, \mu^{i-1}, \sigma^{i-1})$
        \STATE $\bar{f}_{s \rightarrow t}^{i} \leftarrow \text{get-embedding}(f_{s \rightarrow t}^{i})$
        \STATE $\mu^{i} \leftarrow \mu^{i-1} - lr \cdot \nabla_{\mu}\mathcal{L}_{\mu,\sigma}(\bar{f}_{s \rightarrow t}^{i}, \text{TrgEmb})$
        \STATE $\sigma^{i} \leftarrow \sigma^{i-1} - lr \cdot \nabla_{\sigma}\mathcal{L}_{\mu,\sigma}(\bar{f}_{s \rightarrow t}^{i}, \text{TrgEmb})$
    \ENDFOR

    \STATE $(\mu, \sigma) \leftarrow (\mu^N, \sigma^N)$
    \STATE $\mathcal{S}_{s \rightarrow t} \leftarrow \mathcal{S}_{s \rightarrow t} \cup \{(\mu,\sigma)\}$
\ENDFOR
\end{algorithmic}
\end{algorithm}

\subsection{Domain Adaptation of Teacher-Student Model}

After obtaining the transformed features (\( f_{s \rightarrow t} \)) aligned to the target domain semantics using the prompt-based feature alignment mechanism, we inject these adapted features directly into Layer 1 of our teacher model's backbone. Utilizing a small set of stored source-domain images, these transformed features allow the teacher model to quickly fine-tune its detection head for a limited number of epochs, effectively capturing the semantic nuances of the target domain.

Following this brief fine-tuning phase, the adapted teacher model is deployed to perform object detection on initial target-domain instances, generating high-quality predictions. These predictions serve as pseudo-labels to guide the adaptation of the lightweight student model (YOLOv11 nano). Leveraging these pseudo-labels, the student model effectively updates its internal representations, thus becoming accurately adapted to the target domain.

This flexible teacher-student adaptation paradigm ensures that our framework can continually adapt and learn new target-domain distributions without additional constraints. 
\subsection{Captioning Pipeline}
A key component of our proposed framework is the use of textual meta-information to guide domain adaptation by aligning source features with a semantically defined target domain. However, publicly available datasets combining detailed background scenario descriptions with object detection annotations are currently scarce. Existing datasets, such as RSICD~\cite{rsicd}, typically provide textual information tailored for image classification tasks, focusing primarily on the objects present rather than describing the broader environmental scenario. In contrast, our domain adaptation objective explicitly requires detailed textual descriptions encompassing the background context, including location, temporal, and environmental conditions.

To address this gap, we developed an automated caption-generation pipeline using the LLaMA 3.2 Vision Model (11B)~\cite{llama32}. We utilize zero-shot prompting to capture comprehensive background scenario information while excluding descriptions of objects or activities within the scene. Each image was provided to the LLaMA model along with the following structured prompt:

\begin{quote}
\textit{You are to analyze the provided image and describe only the background scenario of the image. Do not include descriptions of objects or ongoing activities. Provide a dictionary output in JSON format with keys 'where', 'when', and 'weather', where 'where' describes the location or exact setting, 'when' refers to the time or lighting condition, and 'weather' captures environmental conditions. Output only the dictionary in JSON format and omit any additional text.}
\end{quote}

Using this structured prompting approach, we systematically generated meta-information for each image, capturing semantic cues for domain adaptation. Given the inherent variability and uncertainty in outputs from large language models, additional processing steps were applied to ensure consistency and reliability across generated captions. Specifically, we standardized terminologies by grouping synonymous or closely related words into unified representations, ensuring a coherent and uniform semantic space across the dataset. Moreover, captions identified as uncertain, ambiguous, or those in which the model's output indicated a lack of confidence were filtered out.

This rigorous caption-generation and refinement process resulted in a high-quality, semantically consistent dataset containing detailed textual descriptions of background scenarios.

\section{Experimental Results and Discussions}

In this section, we present experimental results and discuss the performance of our proposed framework, Prmpt2Adpt, highlighting the experimental setup, dataset division, and domain adaptation scenarios. It is important to note that providing a direct and fair comparison to existing approaches is challenging, given that our method explicitly targets resource-constrained environments. Existing prompt-driven domain adaptation methods generally rely heavily on large-scale VLMs and assume ample storage capacity to retain extensive source-domain datasets. In contrast, our approach deliberately limits the source-domain memory requirement to storing only five representative images, performing adaptation exclusively through these images. Despite this substantial difference in resource assumptions, we provide a  evaluation demonstrating the effectiveness and efficiency of our method under realistic constraints.

\subsection{Dataset}
Our primary experiments are conducted using the MDS-A (Multiple Distribution Shift – Aerial) Dataset \cite{mds_a}. Additionally, we utilize several external datasets—UAVDT~\cite{uavdt}, AID~\cite{AID}, VDD~\cite{VDD}, and NAT2021~\cite{NAT2021}—to fine-tune our distilled CLIP model. To avoid data leakage, we explicitly exclude any images from the MDS-A dataset during this fine-tuning step. Fine-tuning with these diverse datasets ensures our distilled CLIP model learns a robust semantic representation across varied environmental and contextual scenarios.

\textbf{MDS-A Dataset} \cite{mds_a}: The MDS-A dataset consists of synthetic aerial images generated using the AirSim simulator under multiple weather conditions, including clear, rain, snow, fog, dust, and falling leaves. It is designed to benchmark object detection models under controlled distribution shifts, enabling research on robustness, out-of-distribution generalization, and domain adaptation in aerial imagery. The dataset mimics real-world variability by simulating diverse environmental scenarios while maintaining consistent scene structure for fair comparative analysis.

\textbf{UAVDT Dataset} \cite{uavdt}: The UAVDT dataset consists of aerial videos captured in complex urban scenarios. It includes varying lighting conditions, camera perspectives, and vehicle densities, providing valuable diversity for learning robust representations suitable for drone-based urban monitoring applications.

\textbf{AID Dataset} \cite{AID}: The Aerial Image Dataset (AID) is an extensive benchmark for aerial scene classification, comprising 10,000 images across 30 different scene categories. These categories span urban, rural, and natural scenarios, enabling comprehensive fine-tuning across diverse visual contexts.

\textbf{VDD Dataset} \cite{VDD}: The Vehicle and Drone Detection (VDD) dataset consists of aerial images specifically curated for vehicle and drone detection tasks. It provides diverse backgrounds and vehicle scales, facilitating effective representation learning for robust object detection in drone-based applications.

\textbf{NAT2021 Dataset} \cite{NAT2021}: The NAT2021 dataset includes images collected from drone perspectives covering varied natural and artificial landscapes under multiple weather conditions. The dataset provides significant variability in environmental and atmospheric scenarios, aiding the distilled CLIP model in capturing robust semantic and visual relationships across different scenarios.

\subsection{Experimental Setting}
To evaluate the effectiveness of our approach under domain shift, we conduct experiments using the MDS-A dataset, which offers diverse environmental scenarios representative of real-world aerial perception challenges. We designate the clear weather condition as the source domain, while rain, snow, fog, dust, and falling leaves serve as target domains. This controlled setup enables us to systematically analyze how well the model adapts to previously unseen and visually challenging conditions.

While our primary objective is to develop a solution that excels in computational efficiency, memory usage, and adaptation speed—key requirements for resource-constrained environments such as drones—we also assess the detection performance of our method relative to existing state-of-the-art approaches. The experimental results demonstrate that our lightweight framework not only enables rapid, low-overhead adaptation but also achieves competitive mAP performance, validating its practicality and scalability for real-time deployment in dynamic scenarios.

\subsection{Results}
\begin{figure}[!th]
  \centering
  \includegraphics[scale=0.25]{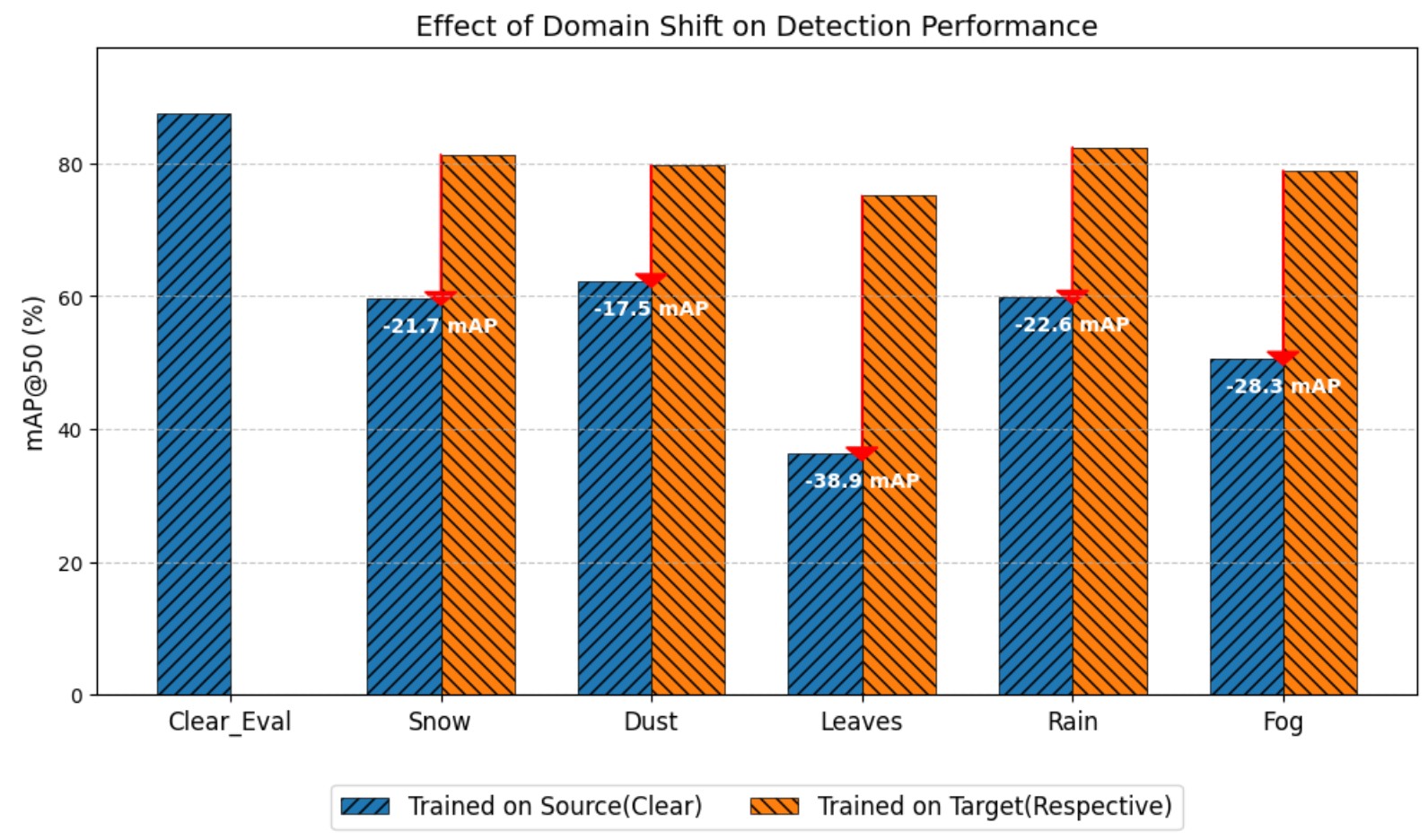}
   \caption{Impact of domain shift on the detection performance of the student model without adaptation. The blue bars represent mAP@50 when the model is trained on the source domain (Clear) and evaluated on each target domain, while the orange bars show performance when the model is trained and tested on the same target domain. Significant performance drops are observed across all target domains, as highlighted by the red arrows, indicating the degradation in detection accuracy due to unseen domain conditions.}

  \label{fig:degrade}
\end{figure}

We first evaluated the baseline performance of the student model (YOLOv11 nano) when trained solely on source domain (clear) and tested directly on the unseen target domains  (rain, snow, fog, dust, and falling leaves). Figure \ref{fig:degrade} summarizes this evaluation, clearly demonstrating substantial performance degradation when the model encounters images from a previously unseen domain. 

Next, we benchmarked our proposed Prmpt2Adpt framework against state-of-the-art prompt-based adaptation methods, PODA and ULDA. For a fair comparison, we adopted the ResNet-50 backbone architecture and implemented analogous detection heads as described in their respective papers. The quantitative results, presented in Table~\ref{tab:result}, show that our lightweight framework achieves competitive detection performance despite its significantly smaller model size and reduced computational requirements.

\begin{figure}[!th]
  \centering
  \includegraphics[scale=0.5]{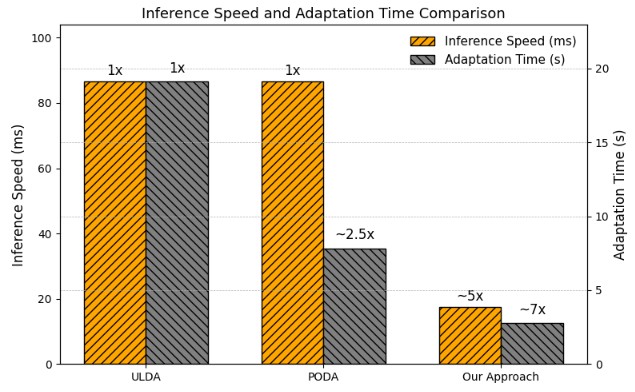}
   \caption{Comparison of inference and adaptation time between our proposed approach and state-of-the-art methods PODA and ULDA. Prmpt2Adpt achieves up to 5× faster inference and 7× faster domain adaptation, demonstrating its efficiency and suitability for real-time deployment in resource-constrained environments. In contrast, PODA and ULDA require significantly more computational time due to their reliance on larger models and full access to source-domain data.}

  \label{fig:speed_up}
\end{figure}

However, the key advantages of our approach become even more evident when examining the efficiency metrics, illustrated in Figure~\ref{fig:speed_up}. While PODA and ULDA require extensive computational resources and adaptation time, our method achieves up to 7× faster domain adaptation and 5× faster inference speed. These gains are largely due to the use of the compact YOLOv11 nano student model and the distilled CLIP-based teacher model, which is half the size of a standard ResNet-50. Collectively, these results demonstrate that Prmpt2Adpt delivers near state-of-the-art accuracy while achieving substantial speedups in both inference and adaptation—making it highly suitable for deployment in time-critical, resource-constrained scenarios.

An additional advantage of our framework lies in its minimal dependency on source-domain data. Unlike PODA and ULDA, which require access to the entire source dataset during adaptation, Prmpt2Adpt operates using only five representative source images, selected and cached prior to deployment. This drastically reduces memory overhead and simplifies data management, further reinforcing the practicality of our method in real-world low-resource settings.

\begin{table}[]
\centering
\caption{Performance Comparison}
\label{tab:result}
\resizebox{210pt}{!}{%
\begin{tabular}{cccccc}
\hline
\multirow{2}{*}{Method}   & \multicolumn{5}{c}{Target   Domain (mAP\%)}                                                                            \\ \cline{2-6} 
                          & \multicolumn{1}{c|}{Snow} & \multicolumn{1}{c|}{Dust} & \multicolumn{1}{c|}{Leaves} & \multicolumn{1}{c|}{Rain} & Fog  \\ \hline
\multicolumn{1}{c|}{PODA \cite{fahes2023poda}} & \multicolumn{1}{c|}{65.7} & \multicolumn{1}{c|}{67.4} & \multicolumn{1}{c|}{44.1}   & \multicolumn{1}{c|}{64.0} & 55.9 \\
\multicolumn{1}{c|}{ULDA \cite{yang2024unified}} & \multicolumn{1}{c|}{67.3} & \multicolumn{1}{c|}{70.1} & \multicolumn{1}{c|}{45.6}   & \multicolumn{1}{c|}{66.2} & 58.1 \\
\multicolumn{1}{c|}{Ours} & \multicolumn{1}{c|}{63.6} & \multicolumn{1}{c|}{64.5} & \multicolumn{1}{c|}{41.2}   & \multicolumn{1}{c|}{62.8} & 53.0 \\ \hline
\end{tabular}%
}
\end{table}


\section{Conclusion}
In this work, we introduced Prmpt2Adpt, a lightweight and prompt-driven framework for zero-shot domain adaptation tailored to resource-constrained environments such as drones. Our approach addresses a fundamental limitation in existing methods, which typically rely on large-scale models and require access to complete source-domain datasets—assumptions that are often infeasible in real-world deployments. By leveraging a distilled and fine-tuned CLIP model for semantic alignment, a teacher-student architecture for efficient knowledge transfer, and a prompt-guided feature steering mechanism, our framework enables fast and robust adaptation to new domains using only a handful of representative source samples and a natural language description of the target domain.

Extensive experiments on the MDS-A dataset demonstrate that Prmpt2Adpt achieves competitive detection performance relative to state-of-the-art models, while significantly outperforming them in terms of computational efficiency, inference speed, and adaptation time. Specifically, our method delivers up to 5× faster inference and 7× faster domain adaptation, making it highly effective for real-time operation under strict resource constraints. In future work, we plan to further improve detection accuracy and extend our framework to support a broader range of domain shifts—including diverse environmental conditions, scenarios, and geographic contexts—to enhance its generalization and real-world applicability.

\section*{Acknowledgements}
This work was supported by the U.S. Military Academy under Cooperative Agreement No. W911NF-22-2-0081. The views and conclusions expressed in this paper are those of the authors and do not reflect the official policy or position of the U.S. Army or U.S. Department of Defense.

{
    \small
    \bibliographystyle{ieeenat_fullname}
    \bibliography{main}
}


\end{document}